\pdfoutput=1

\documentclass[11pt]{article}
\usepackage[table,xcdraw]{xcolor}  
\usepackage{graphicx}
\usepackage{subcaption}

\usepackage[]{ACL2023}

\usepackage{times}
\usepackage{latexsym}
\usepackage{enumitem}
\usepackage[T1]{fontenc}

\usepackage[utf8]{inputenc}

\usepackage{microtype}

\usepackage{inconsolata}

\usepackage{amsmath, amssymb}
\usepackage{tikz}
\usepackage{amsmath}
\usepackage{tcolorbox}
\usepackage{multirow}

\usepackage{amsmath}

\usepackage{colortbl}
\usepackage{multirow} 
\usepackage{booktabs} 
\usepackage{comment}
\usepackage[inkscapelatex=false]{svg}
\usepackage{tcolorbox}
\usepackage{xcolor}
\usepackage{algorithm}
\usepackage{algpseudocode}
\usepackage{afterpage}

\definecolor{lightblue}{rgb}{0.2, 0.4, 0.8} 

\usepackage{array}
\usepackage{tabularx}
\usepackage{arydshln}
\usepackage{float}
\graphicspath{{figures/}}
\definecolor{my-blue}{cmyk}{0.56, 0.07, 0, 0.44, 1.00}
\definecolor{mygreen}{RGB}{92, 214, 92}
\definecolor{myred}{RGB}{255, 92, 51}
\newtcolorbox{boxA}{
    boxrule = 1pt,
    colframe = my-blue,
    colback=white!10!white
}

\definecolor{lightbluetokens}{RGB}{220,230,255}
\definecolor{lightred}{RGB}{255,200,200}
\definecolor{verylightblue}{RGB}{240,245,255}
\definecolor{verylightred}{RGB}{255,230,230}

\title{CURE: Controlled Unlearning for Robust Embeddings --- Mitigating Conceptual Shortcuts in Pre-Trained Language Models}

\author{Aysenur Kocak \quad Shuo Yang \quad Bardh Prenkaj \quad Gjergji Kasneci \\\\
Technical University of Munich \\
\texttt{\{name.surname\}@tum.de}
}

\begin{document}
\maketitle
\begin{abstract}
Pre-trained language models have achieved remarkable success across diverse applications but remain susceptible to spurious, concept-driven correlations that impair robustness and fairness. In this work, we introduce CURE, a novel and lightweight framework that systematically disentangles and suppresses conceptual shortcuts while preserving essential content information. Our method first extracts concept-irrelevant representations via a dedicated content extractor reinforced by a reversal network, ensuring minimal loss of task-relevant information. A subsequent controllable debiasing module employs contrastive learning to finely adjust the influence of residual conceptual cues, enabling the model to either diminish harmful biases or harness beneficial correlations as appropriate for the target task. Evaluated on the IMDB and Yelp datasets using three pre-trained architectures, CURE achieves an absolute improvement of +10 points in F1 score on IMDB and +2 points on Yelp, while introducing minimal computational overhead. Our approach establishes a flexible, unsupervised blueprint for combating conceptual biases, paving the way for more reliable and fair language understanding systems.\footnote{Our code is available at \url{https://github.com/aysenurozmen/CURE}}
\end{abstract}

\section{Introduction}

\begin{figure}[!t]
    \centering
        \begin{boxA}
            \small
            \textcolor{black}{
            \textbf{Training} \\
             - The {\color{lightblue}wood-fired pizza} had the perfect balance of crispy crust, tangy {\color{lightblue}tomato sauce}, and gooey {\color{lightblue}cheese}—absolutely delicious! ($f_\theta$: \textcolor{mygreen}{positive}) \\
             - I wasn’t expecting much, but the homemade {\color{lightblue}lasagna} was rich, and bursting with flavor! ($f_\theta$: \textcolor{mygreen}{positive}) \\
            \rule{\textwidth}{0.2pt}
            \textbf{Testing}\\
             - The {\color{lightblue}breading} was soggy, and the {\color{lightblue}meat} was disappointingly dry. ($f_\theta$: \textcolor{myred}{positive})
            }
    \end{boxA}
    \vspace{-5pt}
    \caption{Example of shortcut learning in sentiment classification, where a classification model $f_\theta$ wrongly associate reviews about \textit{Food} to a \textit{positive} sentiment.}
    \label{fig:example}
    \vspace{-10pt}
\end{figure}

With the rapid advancement of artificial intelligence, pre-trained language models (PLMs) have been widely adopted across various domains, including education, healthcare and e-commerce~\citep{bert, gpt2, llama}. A predominant strategy for applying these models is fine-tuning, where a PLM is further adapted to task-specific data, aiming to enhance its performance or better align with human intent~\citep{instructgpt}.
However, fine-tuning often exposes models to dataset biases, leading to shortcuts—spurious correlations between features and labels~\citep{unlearn}. For instance, \citet{concept} demonstrated that on the Yelp dataset~\citep{yelp}, a LLaMA2-based~\citep{llama2} sentiment classifier mistakenly associated the concept of ``food'' with a ``positive'' label.
These fragile dependencies not only limit the robustness of PLMs but also pose significant risks. In medical diagnosis, a biased detector might incorrectly associate certain biological attributes with diseases, leading to inaccurate predictions~\citep{chest}. Similarly, in automated recruitment systems, a shortcut may result in a favor to applicants with certain demographic attributes, exacerbating fairness problem. In Figure~\ref{fig:example}, we present an example where the classifier incorrectly associates the concept of food with positive sentiment.


Contemporary debiasing research primarily focus on two strategies: (1) modifying shortcut-inducing terms in training data~\citep{cad, razor}, and (2) generating counterfactual samples~\citep{disco, concept} via large language models (LLMs). However, both approaches suffer from notable limitations.
Lexical modification requires prior knowledge of shortcut-inducing terms, which is often challenging to obtain~\citep{much}. Moreover, its effectiveness is restricted to lexical shortcuts rather than conceptual biases.
On the other hand, LLM-based counterfactual generation is computationally expensive and increases training costs significantly. While LLM-free counterfactual generation still relies on prior knowledge~\citep{fever-cf}, making it similarly constrained.

As an unsupervised and lightweight solution, we propose CURE—\underline{C}ontrolled \underline{U}nlearning for \underline{R}obust \underline{E}mbeddings. CURE remaps the semantic space to disentangle conceptual and content-related information without human annotation, offering fine-grained control over shortcut effects. It first trains a content extractor using a concept classifier and back-translation to produce concept-irrelevant representations. A contrastive learning-based debiasing module then refines sample representations, adjusting conceptual features as needed. Finally, the module is jointly trained with a classification head to enhance model robustness.


Unlike traditional approaches, CURE offers three key advantages: 
\textbf{Prior Knowledge Independence} – CURE uses unsupervised learning, eliminating the need for manual annotations of shortcuts.  
\textbf{Resource Efficiency} – 
CURE eliminates the need for LLM-driven data augmentation, reducing the training time to approximately one-tenth of the original.
\textbf{Controllability} – CURE can quantify the impact of conceptual bias on classification results. This facilitates both the mitigation of conceptual biases to enhance performance on out-of-distribution (OOD) data and the exploitation of shortcuts to improve performance on independent and identically distributed (i.i.d.) data. Such adaptability enables users to align training objectives with their generalization requirement, while also providing a quantifiable framework for future debiasing research.

Our contributions are as follows:
\begin{enumerate}
    \item \textbf{We propose a novel conceptual debiasing approach named CURE.} It mitigates shortcuts without relying on prior knowledge or data augmentation, reducing training time to one-tenth of that required by LLM-driven methods. Furthermore, CURE is highly adaptable and can be seamlessly integrated with any mainstream PLM.
    \item \textbf{CURE enables precise control over the impact of shortcuts.} It mitigates conceptual biases to enhance robustness against distribution shifts. Conversely, in scenarios where shortcuts align well with the target task, e.g., i.i.d. data, it leverages them to improve classification accuracy. This adaptability allows CURE to balance robustness and accuracy based on specific generalization requirements.
    
    \item \textbf{We evaluate CURE across two benchmark datasets and three PLMs.} Experimental results indicate that on the IMDB dataset, the RoBERTa-based CURE achieves an approximately 5-point improvement in accuracy over an LLM-driven debiasing approach and outperforms the baseline by about 10 points in F1 score, demonstrating its effectiveness in mitigating conceptual shortcuts.
\end{enumerate}

\section{Related Work}
Addressing spurious correlations in PLMs has become a critical research focus, as these correlations can lead to biased and unreliable predictions, limiting model robustness and fairness. Traditional works have explored various strategies to mitigate these issues, including causal inference techniques~\citep{wang-etal-2022-causal}, adversarial training~\citep{Sagawa*2020Distributionally}, and data augmentation methods designed to reduce model reliance on spurious features~\citep{kaushik-etal-2021-cnlp}. Additionally, approaches leveraging counterfactual reasoning~\citep{fever-cf} have shown promise in improving fairness and robustness in LLMs. These advancements collectively contribute to a growing body of research aimed at developing more reliable and ethically sound language models.
\subsection{General Approaches to Addressing Spurious Correlations}
\citet{kumar2019topics} addresses the challenge of models learning spurious topical shortcuts instead of relevant features in tasks like native language identification. They introduce an adversarial model to demote these latent topical confounds using log-odds ratios, guiding the model to focus on stylistic rather than topic-based features. \citet{yaghoobzadeh2019increasing} enhance robustness by fine-tuning models on ``forgettable'' examples that models initially misclassified. \citet{stacey2020avoiding} tackle the issue of natural language inference (NLI) models relying on superficial hypothesis patterns by using an ensemble of adversarial classifiers. \citet{wang2020identifying} propose using treatment effect estimation to distinguish genuine correlations from spurious ones, such as associating ``Spielberg'' with positive sentiment in movie reviews. \citet{wang2021identifying} extend this concept with an automated framework using interpretability techniques, cross-dataset stability, and knowledge-aware perturbation to identify spurious tokens at scale. \citet{tu2020empirical} explores how pre-trained models like BERT handle spurious correlations, finding that they improve robustness by generalizing from minority counterexamples. The authors propose using multi-task learning (MTL) with auxiliary tasks to enhance robustness when these counterexamples are scarce. \citet{du2022less} propose the Less-Learn-Shortcut (LLS) which down-weighs examples with high correlations between specific words and labels. \citet{fever-cf} present a counterfactual debiasing approach that balances predictions between claim-only and claim-evidence models to reduce bias associated with claim patterns.
While these studies primarily address general spurious correlations, recent research has started focusing on spurious correlations at the concept level.

\subsection{Concept-Level Spurious Correlations}

\citet{concept} introduce biases in NLP at the concept level, highlighting how language models often rely on broad associative patterns rather than deeper semantic understanding. For instance, models may learn to associate certain concepts, such as “food”, with inherently positive sentiment, leading to spurious correlations that degrade generalization performance. To mitigate this issue, the authors leverage LLM to generate counterfactual data that rebalances label distributions, thereby reducing the reliance on such superficial cues. 

However, this approach presents certain limitations in terms of scalability. Specifically, generating counterfactual data for each new task requires substantial manual intervention, as it involves defining relevant concept-level biases and ensuring the generated data maintains both linguistic plausibility and task relevance. Even with advanced LLMs like ChatGPT, this process remains resource-intensive, particularly for large-scale or multi-domain applications. Additionally, the effectiveness of this method depends on the quality and diversity of the generated counterfactuals, which can vary depending on the prompt design and the inherent biases present in the language model used for data generation.
These challenges underscore the need for more automated, generalizable approaches to mitigating concept-level biases in NLP.

\section{Methodology}
\begin{figure*}[!t]
        \centering     \includegraphics[width=\textwidth]{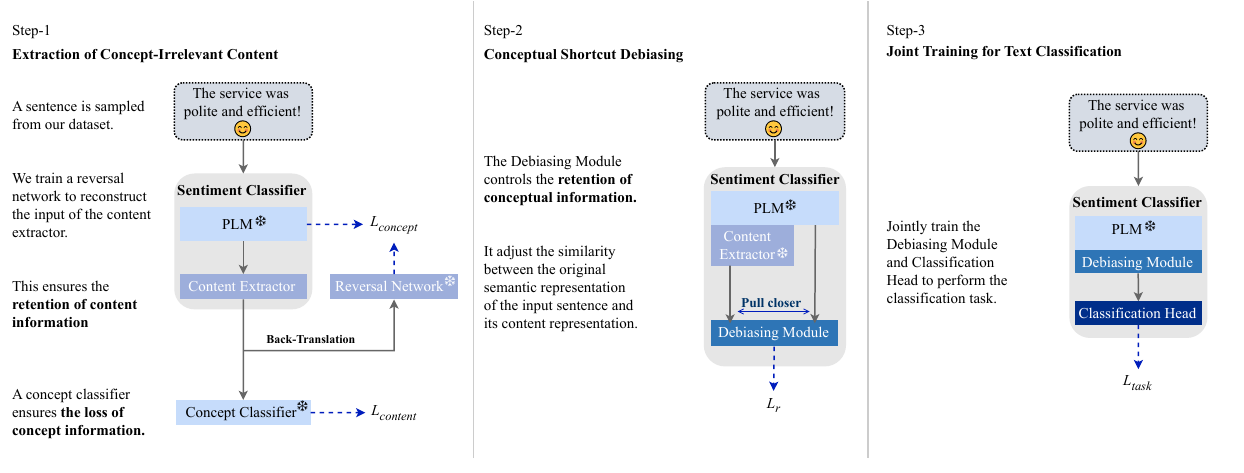}
            \caption{The training process of our CURE involves three steps: 1) We train a content extractor to filter out concepts while retaining concept-irrelevant information using a reversal network. 2) The PLM outputs are remapped through a debiasing network, regulating concept retention by controlling the relationship between the original and content representations. 3) We jointly train the classification head and the debiasing network to maximize robust feature retention while filtering out conceptual information. ``\textasteriskcentered'' indicates frozen model parameters.}
            \label{fig:overview}
\end{figure*}

\subsection{Problem Formulation}

Given a set of i.i.d. labeled documents – $D = \{d_1, \ldots, d_N\}$, where each sample $d_i$ associates with a conceptual label $c_i \in \mathcal{C}$ and a classification label $y_i \in \mathcal{Y}$. We assume that the classification labels are balanced, while the conceptual labels are biased. That is, for every label $y \in \mathcal{Y}$, the number of samples in $D$ with label $y$ is equal:
\begin{equation}
\forall\, y \in \mathcal{Y}, \quad \left| \{ d_i \in D \mid y_i = y \} \right| = \frac{N}{|\mathcal{Y}|}.
\end{equation}
The distribution of conceptual labels is uneven:
\begin{equation}
\begin{gathered}
    \exists\, c, c' \in \mathcal{C} \text{ such that }\\
    \left| \{ d_i \in D \mid c_i = c \} \right| \neq \left| \{ d_i \in D \mid c_i = c' \} \right|.
\end{gathered}
\end{equation}
Here, $\mathcal{C}$ is correlated with but is not causal related to $\mathcal{Y}$, i.e., $\mathcal{C}\perp \!\!\!\perp\mathcal{Y}, \text{ but } \mathcal{C}\nrightarrow\mathcal{Y}$.
We first transform samples in $D$ to their semantic embedding $\mathcal{X}= \{x_1, \ldots, x_N\}\subseteq\mathbb{R}^u$ by using a PLM, then optimize a classification head $f_\theta$ with parameter $\theta$ for mapping $\mathcal{X}\rightarrow\mathcal{Y}$ by minimizing classification loss $\ell$:
\begin{equation}
\theta^* = \arg\min_{\theta} \frac{1}{n} \sum_{i=1}^{n} \ell \big( f_\theta(x_i), y_i \big).
\label{eq:ce}
\end{equation}%
However, due to the bias between $\mathcal{Y}$ and $\mathcal{C}$, the model may erroneously associate $c_i$ with $y_i$, thereby losing its robustness. Our primary objective is to enhance the robustness of $f_\theta$, measured by its classification accuracy on a conceptually balanced OOD test set. 

\subsection{Concept Labeling}\label{subsec:labeling}
Due to the lack of available conceptual annotations in classification datasets and the demonstrated capability of LLMs to perform text annotation~\citep{Gilardi2023ChatGPTOC}, we employ the standard text conceptual annotation pipeline outlined in \citep{concept} by using GPT-4o~\citep{instructgpt}. 

Specifically, we preprocess $D$ with the following three steps: 
\begin{enumerate}
    \item \textbf{Data Cleaning:} We remove uninformative content, including non-ASCII characters and irrelevant metadata from texts. 
    \item \textbf{Concept Labeling:} We design structured prompts (see Appendix~\ref{prompt}) and input them into GPT-4o to label each sample $d_i$ with a concept $c_i$.
    \item \textbf{Meta-Concept Merging:} The generated concepts are then automatically categorized and merged by GPT-4o into a meta-concept set $\mathcal{C}$. 
\end{enumerate}

After obtaining the concept set $\mathcal{C}$, we compute the mutual information between a concept $c$ and $\mathcal{Y}$ to quantify the bias of a specific concept:
\begin{equation}
I(c; \mathcal{Y}) =  \sum_{y \in \mathcal{Y}} P(c, y) \log \frac{P(c, y)}{P(c) P(y)}.
\end{equation}%
Subsequently, we select samples with the top $k$ concepts with the highest $I(c; \mathcal{Y})$ as the training set for training a biased benchmark. Furthermore, we treat samples with the $k$ concepts with the lowest mutual information as the OOD data from real world to evaluate our debiasing method.

\subsection{Extraction of Concept-Irrelevant Content}

To mitigate the impact of conceptual biases, we first extract concept-irrelevant content representations from a semantic embedding $x$. To achieve this, we freeze the parameters of the PLM and insert a lightweight network $f_\phi$ to its output layer, as a content extractor. Here, our objective is to maximize the dropout of concept-related features, while maximizing the retention of content-related features. Therefore, the training loss consists of two components: \textbf{a concept dropout loss}, and \textbf{a content retention loss}.

\subsubsection{Conceptual Information Filter}

We first train a concept classifier to quantify the retention of concept-related features in $\mathcal{X}$. This classifier consists of a classification head $f_\omega$, built on the same PLM as the task classifier $f_\theta$. We optimize parameter $\omega$ by maximizing the probability for predicting $\mathcal{C}$ from $\mathcal{X}$:
\begin{equation}
\omega^* = \arg\min_{\omega} \frac{1}{N} \sum_{i=1}^{N} \ell \big( f_\omega(x_i), c_i \big),
\label{eq:concept}
\end{equation}
where $\ell$ is the cross-entropy loss, defined as:
\begin{equation}
\ell \big( f_\omega(x_i), c_i \big) = - \log P(c_i\mid x_i; \omega),
\end{equation}
where $P(c_i \mid x_i; \omega)$ denotes the predicted probability of concept $c_i$ given input $x_i$, obtained from the softmax output of $f_\omega$. 

We expect the conceptual information in $x$ to be filtered out after transformation by the content extraction function $f_\phi$. To enforce this constraint, we compute the Kullback-Leibler (KL) divergence between the predicted distribution of the concept classifier $\omega$ and a uniform distribution over $\mathcal{C}$ as the training loss $\mathcal{L_\text{concept}}(\phi)$, as shown in Eq.~\eqref{eq:kl_loss}. 

\begin{equation}
\sum_{c \in \mathcal{C}} P(c \mid f_\phi(x); \omega) \log \left( \frac{P(c \mid f_\phi(x); \omega)}{(1 / |\mathcal{C}|)^\tau} \right),
\label{eq:kl_loss}
\end{equation}
where $\tau$ is a temperature parameter that controls the strength of the distribution alignment.

With the training of $f_\phi$, we force the concept classifier $\omega$ to produce maximally uncertain predictions, indicating the absence of learnable conceptual information.

\subsubsection{Concept-Irrelevant Content Maintenance}

The semantic features are often entangled with each other~\citep{styletransformer}. As a result, although the content extractor $f_\phi$ solely aims at filtering out conceptual information, it is crucial to ensure that the concept-irrelevant information remain complete. Inspired by back-translation in machine translation~\citep{backtrans}, we construct a reversal network $\hat{\phi}$, with the same architecture as $f_\phi$. $\hat{\phi}$ is designed to reconstruct $x$ from $f_\phi(x)$, ensuring that the mapping function $f_\phi$ does not excessively lose the concept-irrelevant information. We first freeze $\phi$, then use the following loss to train $\hat{\phi}$:
\begin{equation}
\mathcal{L}(\hat{\phi}) = \left\| f_{\hat{\phi}}\left(f_\phi(x)\right) - x \right\|_2^2.
\label{eq:reconstruction_loss}
\end{equation}%

Next, we freeze the parameters of $\hat{\phi}$. During the training of $\phi$, we use it to remap $f_\phi(x)$ back to $x$, and compute the mean squared error between them as a content retention loss:
\begin{equation}
\mathcal{L}_{\text{content}}(\phi) = \| f_{\hat{\phi}}(f_\phi(x)) - x \|^2_2.
\end{equation}%
Finally, we combine $\mathcal{L}_{\text{content}}$ and $\mathcal{L}_{\text{concept}}$ with weighted summation to form the overall loss for training $f_\phi$, as shown in eq.~\eqref{eq:phi}. 
\begin{equation}
\mathcal{L}(\phi) = \mathcal{L}_{\text{concept}}(\phi) + \lambda \mathcal{L}_{\text{content}}(\phi),
\label{eq:phi}
\end{equation}
where $\lambda$ is the weighing to control the relative importance of the content retention.

Finally, we alternately train $f_\phi$ and $f_{\hat{\phi}}$ to ensure that $f_{\hat{\phi}}$ can effectively track the retention of concept-irrelevant information by $f_\phi$. Here, $\phi$ and $\hat{\phi}$ minimizes conceptual information while maximizing the content information, forming an information bottleneck~\citep{tishby99information}.

After training, the classifier $\phi$ maximizes the retention of content while minimizing the retention of conceptual information to avoid conceptual shortcut in further training. Here, $f_\phi(x)$ is the content representation of $x$, denoted as $x_\text{cont}$.

\subsection{Conceptual Shortcut Debiasing}
Although $x_{\text{cont}}$ can replace the original embedding $x$ to mitigate the conceptual bias, we further argue that eliminating conceptual shortcuts is not always beneficial. Theoretically, we identify two special cases where preserving conceptual biases could be advantageous: (1) when the conceptual bias aligns with human intent, and (2) when the application scenario is constrained, where the optimization objective is limited to i.i.d. data.

When the imbalance of conceptual attributes aligns with human natural intent, the shortcuts should be enhanced. For example, in the movie review dataset IMDB~\citep{imdb}, most reviews labeled by GPT-4o as containing the conceptual attribute of ``humor'' are positive. This observation is consistent with psychological studies on relation between language styles and sentiments, which suggest that humorous expression tends to be associated with positive emotion~\citep{humor}. 
Furthermore, for certain application scenario where the i.i.d. and OOD data distribution is identical, the real-world data hold the same distributional bias. For instance, in clinical medicine, a model trained on electronic health records collected from a specific hospital is often deployed to the same environment~\citep{pmlr-v174-hur22a}, classifying text with similar biases in training and inference. In such a case, reinforcing shortcuts can also improve classification performance in application.

To achieve flexible control over the shortcut exploitation, we introduce a lightweight feedforward network $\psi$ on top of the frozen content extractor $f_\phi$ and the PLM. This network maps both the original embedding $x$ and its content representation $x_{\text{cont}}$ into a conceptually controlled semantic space $\mathcal{X_\text{CURE}}\subseteq\mathbb{R}^u$. We then employ contrastive learning to regulate their cosine similarity in this space. The training losses for removing the conceptual shortcut $\mathcal{L}_{\text{r}}(\psi)$ and enhancing the shortcut $\mathcal{L}_{\text{e}}(\psi)$ as follows:

\begin{equation}
 \mathcal{L}_{\text{r}}=\max \left( 0, 1-\cos(f_\psi{(x)}, f_\psi{(x_{\text{cont}})})-\text{M} \right),
 \label{eq:margin1}
\end{equation}

\begin{equation}
 \mathcal{L}_{\text{e}}=\max \left( 0, \cos(f_\psi{(x)}, f_\psi{(x_{\text{cont}})})-\text{M} \right),
  \label{eq:margin2}
\end{equation}
where $\text{m}\in[0,1]$ is a margin that controls the degree of conceptual information retention. 

A smaller margin $\text{M}$ enforces a stricter optimization objective. In the removal loss $\mathcal{L}_{\text{r}}$, decreasing $\text{M}$ compels $f_\psi{(x)}$ and $f_\psi{(x_{\text{cont}})}$ to be nearly identical, ensuring the complete removal of conceptual information. Conversely, in the enhancement loss $\mathcal{L}_{\text{e}}$, a smaller $\text{M}$ forces $f_\psi{(x_{\text{cont}})}$ and $x_{\text{cont}}$ to be maximally separated, thereby amplifying the influence of conceptual features. By adjusting $\text{M}$, we can flexibly control the extent to which conceptual information is retained in $f_\psi{(x)}$.

Finally, we replace the original embedding $x$ with $f_\psi{(x)}$, as the input to the classifier $f_\theta$ and jointly train $f_\theta$ and $f_\psi$ using Equation~\eqref{eq:ce}. The trained model can flexibly adjust the extent of conceptual bias retention based on the training objective, making it either more robust or more \textit{specialized}, as shown in Fig.~\ref{fig:overview}. 
In terms of parameter efficiency, CURE introduces only a lightweight content extractor and feedforward network on top of the original classifier, ensuring minimal computational overhead.

\begin{table*}[!t]
\centering
\resizebox{\textwidth}{!}{
\begin{tabular}{lllllllllllllll}
\toprule
\multicolumn{2}{c}{Dataset}  & \multicolumn{6}{c}{\textbf{IMDB}}    & \multicolumn{1}{c}{}     & \multicolumn{6}{c}{\textbf{Yelp}}     \\ \cline{3-8} \cline{10-15} 
\multicolumn{2}{c}{Model}    & \multicolumn{2}{c}{DistilBERT}    & \multicolumn{2}{c}{MPNet}      & \multicolumn{2}{c}{RoBERTa}       & \multicolumn{1}{c}{}     & \multicolumn{2}{c}{DistilBERT}   & \multicolumn{2}{c}{MPNet}      & \multicolumn{2}{c}{RoBERTa}    \\
&    & \multicolumn{1}{c}{ACC $\uparrow$}                & \multicolumn{1}{c}{F1 $\uparrow$}                 & \multicolumn{1}{c}{ACC $\uparrow$}                & \multicolumn{1}{c}{F1 $\uparrow$}                 & \multicolumn{1}{c}{ACC $\uparrow$}                & \multicolumn{1}{c}{F1 $\uparrow$}                 & \multicolumn{1}{c}{}     & \multicolumn{1}{c}{ACC $\uparrow$}                & \multicolumn{1}{c}{F1 $\uparrow$}                 & \multicolumn{1}{c}{ACC $\uparrow$}                & \multicolumn{1}{c}{F1 $\uparrow$}                 & \multicolumn{1}{c}{ACC $\uparrow$}                & \multicolumn{1}{c}{F1 $\uparrow$}                 \\ \hline
                         & Baseline                     & \underline{84.00}                            & \underline{85.05}                            & 87.33                                  & 86.94                                  & 88.50                                  & \underline{89.27}                            &                          & 94.75                                  & 94.76                                  & 92.75                                  & 93.11                                  & 93.75                                  & 93.51                                  \\
                         & FL                           & 83.70                                  & 82.00                                  & \underline{87.50}                            & \underline{87.32}                            & \underline{88.67}                            & 88.90                                  &                          & 92.25                                  & 92.54                                  & 93.75                                  & \textbf{95.17}                         & \underline{93.50}                            & \underline{93.00}                            \\
                         & RAZOR                        & 83.25                                  & 83.00                                  & 87.00                                  & 86.50                                  & 85.33                                  & 83.19                                  &                          & \textbf{95.50}                         & \textbf{95.32}                         & \underline{93.50}                            & 94.83                                  & 92.50                                  & \underline{93.00}                            \\
\multirow{-4}{*}{i.i.d.} & \cellcolor[HTML]{EFEFEF}CURE & \cellcolor[HTML]{EFEFEF}\textbf{85.50} & \cellcolor[HTML]{EFEFEF}\textbf{85.48} & \cellcolor[HTML]{EFEFEF}\textbf{88.83} & \cellcolor[HTML]{EFEFEF}\textbf{88.78} & \cellcolor[HTML]{EFEFEF}\textbf{89.67} & \cellcolor[HTML]{EFEFEF}\textbf{89.77} & \cellcolor[HTML]{EFEFEF} & \cellcolor[HTML]{EFEFEF}\underline{95.25}    & \cellcolor[HTML]{EFEFEF}\underline{95.25}    & \cellcolor[HTML]{EFEFEF}\textbf{95.00} & \cellcolor[HTML]{EFEFEF}\underline{95.00}    & \cellcolor[HTML]{EFEFEF}\textbf{94.75} & \cellcolor[HTML]{EFEFEF}\textbf{94.63} \\ \hline
                         & Baseline                     & \underline{81.67}                            & 82.20                                  & \underline{79.33}                            & \underline{80.19}                            & 78.83                                  & 74.85                                  &                          & 89.75                                  & 90.44                                  & 89.00                                  & 88.30                                  & 89.25                                  & 89.64                                  \\
                         & FL                           & 81.33                                  & \underline{82.25}                            & 79.00                                  & 76.75                                  & \underline{79.33}                            & 76.70                                  &                          & 90.25                                  & 89.53                                  & 90.25                                  & \underline{89.40}                            & 89.00                                  & 89.52                                  \\
                         & RAZOR                        & 80.83                                  & 81.30                                  & 79.00                                  & 79.33                                  & 78.67                                  & \underline{77.70}                            &                          & \underline{90.75}                            & \underline{90.60}                            & \textbf{90.75}                         & 89.26                                  & \underline{89.50}                            & \underline{89.76}                            \\
\multirow{-4}{*}{OOD}    & \cellcolor[HTML]{EFEFEF}CURE & \cellcolor[HTML]{EFEFEF}\textbf{84.00} & \cellcolor[HTML]{EFEFEF}\textbf{84.36} & \cellcolor[HTML]{EFEFEF}\textbf{81.50} & \cellcolor[HTML]{EFEFEF}\textbf{81.22} & \cellcolor[HTML]{EFEFEF}\textbf{83.50} & \cellcolor[HTML]{EFEFEF}\textbf{84.51} & \cellcolor[HTML]{EFEFEF} & \cellcolor[HTML]{EFEFEF}\textbf{92.00} & \cellcolor[HTML]{EFEFEF}\textbf{92.12} & \cellcolor[HTML]{EFEFEF}\textbf{90.75}    & \cellcolor[HTML]{EFEFEF}\textbf{90.68} & \cellcolor[HTML]{EFEFEF}\textbf{91.50} & \cellcolor[HTML]{EFEFEF}\textbf{91.33} \\ 
\bottomrule
\end{tabular}
}
\caption{Accuracy and F1 on i.i.d. and OOD test on the IMDB and Yelp datasets. ``Baseline'' stands for PLMs fine-tuned solely on classification tasks. ``ACC'' stands for Accuracy; \textbf{Bolded} values indicate best performing; \underline{underlined} the second-best.}
\label{tab:main}
\vspace{-5pt}
\end{table*}

\section{Experiments}

\subsection{Experimental Setup}

\paragraph{Dataset Description}
We used IMDB~\citep{imdb} and Yelp~\citep{yelp} datasets. The IMDB movie review dataset is a binary sentiment analysis dataset, which consists of 50,000 positive or negative reviews from the Internet Movie Database. The Yelp dataset is provided by the Yelp Dataset Challenge, contains business reviews labeled with ratings ranging from 0 to 4~\citep{yelp}. We used the version that was cleaned and organized by~\citet{styletransformer}.

Based on the concepts labeled in Section~\ref{subsec:labeling}, we divided the samples in the each dataset into two groups for i.i.d. and OOD testing:
\begin{itemize}
    \item \textbf{Group A} contains \textbf{imbalanced} concept distributions, where certain concepts are overrepresented in one task-relevant category, but the overall number of samples across task-relevant labels remains equal. Samples in Group A will be separate to a biased training set and an i.i.d. test set.
    \item \textbf{Group B} contains \textbf{balanced} concept distributions, where each concept has an equal number of samples across the task-relevant categories. Samples in Group B will be used as the OOD test set.
\end{itemize}

\paragraph{Compared Methods and Hyperparameters}
As there are currently no model-based debiasing approaches, we primarily compare our method with FL~\citep{focalloss}, which optimizes loss computation with unbalanced data, and RAZOR~\citep{razor}, which utilizes LLMs for data debiasing. The result is shown in Table~\ref{tab:main}.

In our training, we used a mini-batch size of 16, with the optimizer AdamW~\citep{adamw}. The learning rate for the content extractor and reversal extractor was set to 0.0001, while that for the classification heads was set to 0.0003. The concept classifier head and task classifier head have identical structures and are based on the same PLM.

\paragraph{Computational Efficiency}
CURE is highly lightweight. Specifically, the content extractor used consists of two single linear layers with layer normalization and a single Transformer layer~\citep{bert}, each with 768 neurons, resulting in a total of approximately 1.78M parameters. Our debiasing module comprises a SwiGLU layer~\citep{swiglu} followed by a linear layer, with a total of approximately 1.18M parameters. We make a comparison with GPT-3.5-Turbo-based RAZOR in Table~\ref{tab:cost}. Here, we calculated the average training and inference time per sample with a batch size of 16 on a single NVIDIA A100 Tensor Core-Graphics Processing Unit.

\paragraph{Case Study} To better understand CURE’s improvements, we analyze the model's attention patterns in sentiment classification tasks. 
Specifically, we randomly sampled a positive review from Yelp, using sentiment classifiers based on DistilBERT and CURE to classify it. After that, we studied their attention across different terms, which is measured by Shapley~\citep{shap}. The attribution visualizations in Table~\ref{tab:baseline} and Table~\ref{tab:cure} highlight these differences.

\begin{table}[ht]
\resizebox{\columnwidth}{!}{
\begin{tabular}{lllc}
\toprule

    Model    & \multicolumn{1}{c}{Scale $\downarrow$} & \multicolumn{1}{c}{Training $\downarrow$}        & \multicolumn{1}{c}{Inference $\downarrow$}      \\ \hline
RoBERTa & 125M                & $\approx$ 11ms & $\approx$ 1ms \\
RAZOR   & GPT-3.5-Turbo  & $>$ 600ms         & $\approx$ 1ms \\
\rowcolor[HTML]{EFEFEF} 
CURE    & 127.96M             & $\approx$ 59ms & $\approx$ 1ms \\ 
\bottomrule
\end{tabular}}
\caption{Computational scale and the average training/inference time per sample. We take the Yelp dataset with RoBERTa as an example.}
\label{tab:cost}
\vspace{-5pt}
\end{table}

\begin{figure}[t]
      \centering
      \begin{minipage}{0.493\linewidth}
        \centering
        \includegraphics[width=\linewidth]{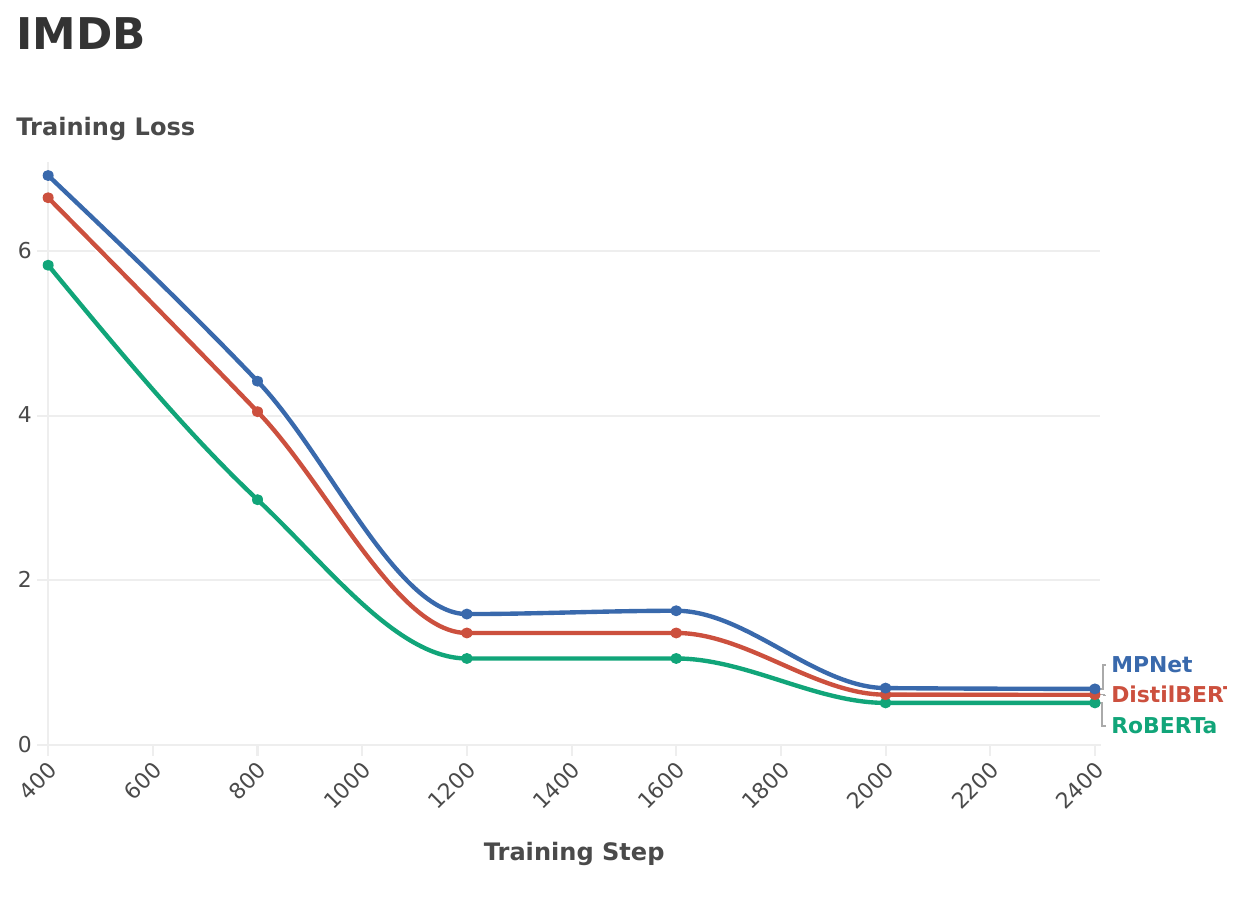}
      \end{minipage}\hfill
      \begin{minipage}{0.493\linewidth}
        \centering
        \includegraphics[width=\linewidth]{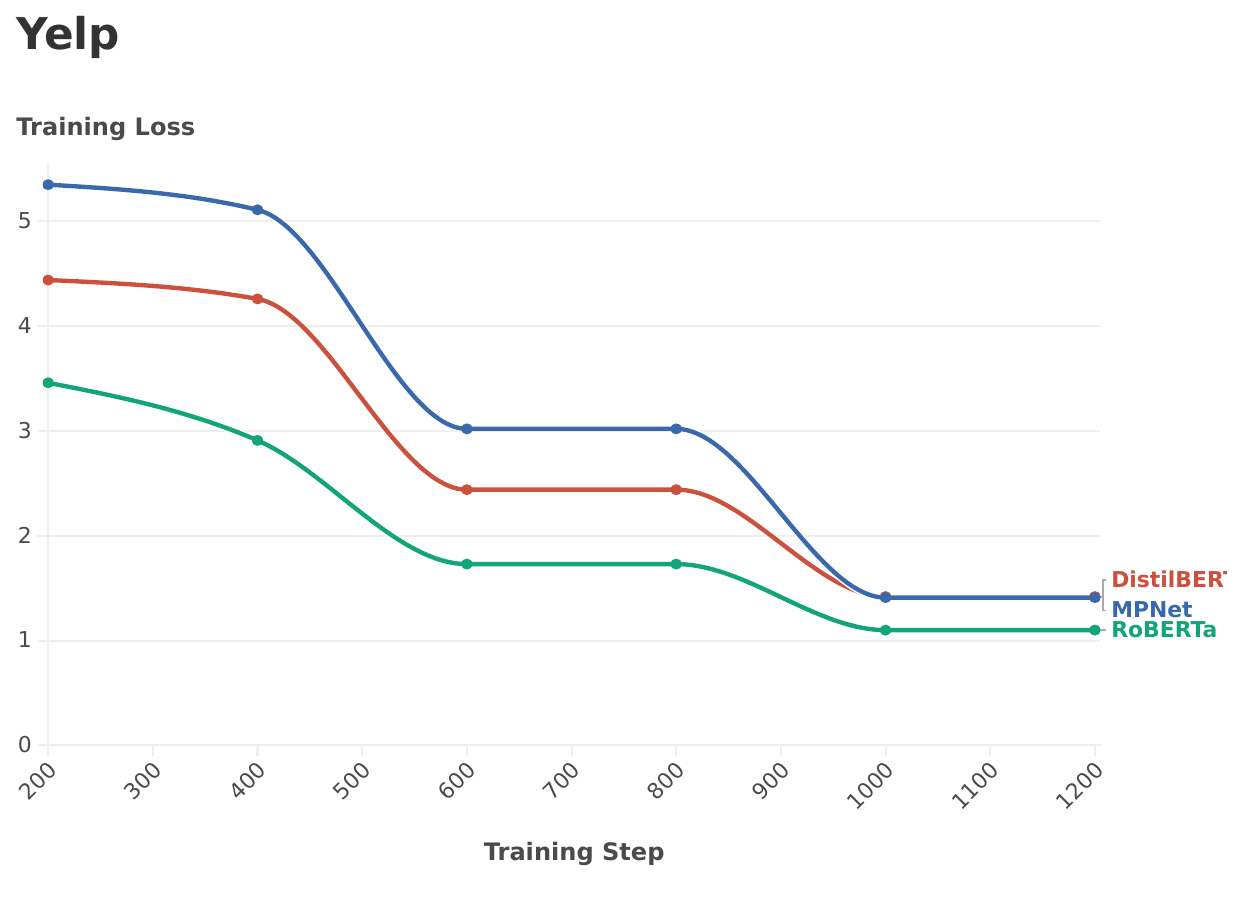}
      \end{minipage}
    
      \caption{The convergence of the content extractor. We scale the loss values by a factor of 100 for clear comparison.}
      \label{fig:convergence}
      \vspace{-5pt}
    \end{figure}

\paragraph{The Convergence of the Content Extractor} Since the content extractor $\phi$ is optimized by two training objectives simultaneously, i.e., $\mathcal{L}_{\text{content}}(\phi)$ and $\mathcal{L}_{\text{concept}}(\phi)$, we empirically demonstrated its convergence. The training curve of the content extractor is shown in Fig.~\ref{fig:convergence}.

\subsection{Results and Discussions}
CURE outperformed the baselines on nearly all metrics across both datasets. as shown in Table~\ref{tab:main}. The largest improvement comes from the Roberta model on the IMDB with the OOD test, with an approximate increase of 5 points in Accuracy and 10 points in F1 score. 
Compared to the i.i.d test, our model introduced a more significant improvement on the OOD test. We analyze that the benchmarks on the i.i.d. test have achieved relatively high accuracy, making it challenging to further improve their performances. Furthermore, we observe that CURE outperforms loss adjustment method FL and LLM-driven approach RAZOR. We attribute this to the fact that FL and RAZOR primarily address label- and word-level biases rather than conceptual biases. For semantic-level biases, these two methods lack mechanisms for regulating the semantic representations, making it challenging for them to improve the baselines. In contrast, CURE remaps the semantic space, enabling the controllable filtering of concept information that cause shortcuts, thereby enhancing  robustness of baselines and boosting their OOD performances.

Our findings show that the baseline model tends to distribute attention across both sentiment-related and domain-specific words, while CURE prioritizes sentiment-expressive terms.  Table~\ref{tab:baseline} illustrates how the DistilBERT-based classifier assigns nearly equal importance to both “service” and “great”, which indicates a reliance on topic-specific terms rather than sentiment indicators. In contrast, Table~\ref{tab:cure} shows that CURE places stronger emphasis on “great”, which suggests it better captures the actual sentiment while reducing confounding biases.

\begin{table}[ht]
    \centering
    \renewcommand{\arraystretch}{1.5} 
    \setlength{\tabcolsep}{4pt} 
    \resizebox{\columnwidth}{!}{
    \begin{tabular}{|c|c|c|c|c|c|c|}
        \hline
        \cellcolor{lightbluetokens} \textbf{[CLS]} & 
        \cellcolor{lightbluetokens} the & 
        \cellcolor{lightred} service & 
        \cellcolor{lightbluetokens} was & 
        \cellcolor{lightred} great & 
        \cellcolor{lightbluetokens} . & 
        \cellcolor{lightbluetokens} \textbf{[SEP]} \\ 
        \hline
    \end{tabular}}
    \caption{SHAP-based token attribution visualization - DistilBERT. \textcolor{red}{Red} represents the contribution to positive sentiment. ``[CLS]'' and ``[SEP]'' are special tokens.}
    \label{tab:baseline}
    \vspace{-10pt}
\end{table}

\vspace{1cm} 

\begin{table}[ht]
    \centering
    \renewcommand{\arraystretch}{1.5} 
    \setlength{\tabcolsep}{4pt} 
    \resizebox{\columnwidth}{!}{
    \begin{tabular}{|c|c|c|c|c|c|c|}
        \hline
        \cellcolor{lightbluetokens} \textbf{[CLS]} & 
        \cellcolor{lightbluetokens} the & 
        \cellcolor{verylightblue} service & 
        \cellcolor{lightbluetokens} was & 
        \cellcolor{red!30} great & 
        \cellcolor{lightbluetokens} . & 
        \cellcolor{lightbluetokens} \textbf{[SEP]} \\ 
        \hline
    \end{tabular}}
    \caption{SHAP-based token attribution visualization - CURE. \textcolor{red}{Red} represents the contribution to positive sentiment. ``[CLS]'' and ``[SEP]'' are special tokens.}
    \label{tab:cure}
\end{table}

CURE is lightweight and efficient, as shown in Table~\ref{tab:cost}. Compared to the baseline, CURE holds only 2\% additional parameters with a nearly identical inference time. Compared to RAZOR which is based on GPT-3.5-Turbo, CURE does not require participating of LLMs during training, which reduces the training time to approximately one-tenth of RAZOR’s. Additionally, the time complexity of the debiasing module involved in inference is $\mathcal{O}(L \cdot H^2)$, where $L$ represents the input length and $H$ denotes the hidden state dimension, which aligns with that of the PLMs used~\citep{attention}. Therefore, the usage of CURE does not alter the time complexity of the baselines. This substantially reduces both computational and time costs that enhances the practicality and generalizability of CURE in real-world applications.

The content extractor used can converge under all conditions, as shown in Fig.~\ref{fig:convergence}. This not only provides an experimental foundation for CURE but also indicates that the two optimization objectives employed, i.e. $\mathcal{L}_{\text{content}}(\phi)$ and $\mathcal{L}_{\text{concept}}(\phi)$, are not in conflict. We argue that this finding supports that concept information is not entirely tangled with the semantic information in the latent space, thereby offering a theoretical basis for future work on feature disentanglement.

\subsection{Ablation Study}
\paragraph{The Effectiveness of Back-Translation} To investigate the effect of the reversal network used in training, we conducted ablation experiments on the reversal network, as shown in Table~\ref{tab:ablationA}.

\begin{table}[ht]

\resizebox{\columnwidth}{!}{
\begin{tabular}{lccccc}
\toprule
                             & \multicolumn{2}{c}{Yelp} &  & \multicolumn{2}{c}{IMDB} \\ \cmidrule{2-3} \cmidrule{5-6} 
\multirow{-2}{*}{}    & ACC $\uparrow$         & F1  $\uparrow$        &  & ACC $\uparrow$         & F1  $\uparrow$        \\ \hline
RoBERTa(w/o $\hat{\phi}$)    & 79.75       & 83.09      &  & 81.33       & 79.03      \\
\rowcolor[HTML]{EFEFEF} 
RoBERTa(w/ $\hat{\phi}$)     & 91.50       & 91.33      &  & 83.50       & 84.51      \\\midrule
MPNet(w/o $\hat{\phi}$)      & 90.25       & 89.71      &  & 79.83       & 78.73      \\
\rowcolor[HTML]{EFEFEF} 
MPNet(w/ $\hat{\phi}$)       & 90.75       & 90.68      &  & 81.50       & 81.22      \\\midrule
DistilBERT(w/o $\hat{\phi}$) & 91.50       & 91.05      &  & 80.83       & 82.12      \\
\rowcolor[HTML]{EFEFEF} 
DistilBERT(w/ $\hat{\phi}$)  & 92.00       & 92.12      &  & 84.00       & 84.36      \\ 
\bottomrule
\end{tabular}}
\caption{Ablation study on the reversal network $\hat{\phi}$. ``w/'' and ``w/o'' represent ``with'' and ``without'', respectively.}
\label{tab:ablationA}

\end{table}

We found that removing the reversal network results in a degradation in classification accuracy, as shown in Table~\ref{tab:ablationA}. The most significant decline was observed with the RoBERTa model on the Yelp dataset, with a decrease of approximately 12 points in accuracy and 8 points in F1 score. Our further experiments revealed that the content extractor exhibited parameter sparsity in the absence of the reversal network. 

Based on these observations, we hypothesize that, without control of content preservation, the content extractor attempts to map all inputs to similar representations, causing its output to become indistinguishable by the concept classifier and leading to the minimization of the loss $\mathcal{L}_\text{concept}$. In such a case, due to the information loss on robust features, the classifiers struggled to obtain sufficient effective features for learning, leading to a decline in performance.

\begin{figure}[t]
  \centering
  \begin{subfigure}{0.493\linewidth}
    \centering
    \includegraphics[width=\linewidth]{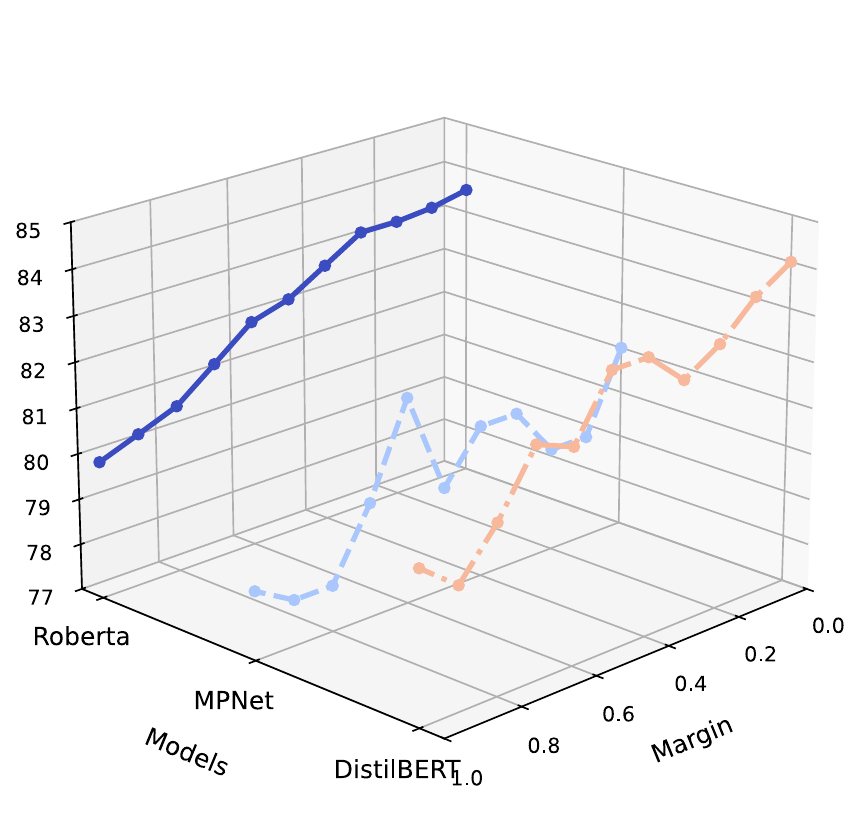}
    \caption{IMDB (Debiasing)}
    \label{fig:de_imdb}
  \end{subfigure}
  \hspace{-0.2em}
  \begin{subfigure}{0.493\linewidth}
    \centering
    \includegraphics[width=\linewidth]{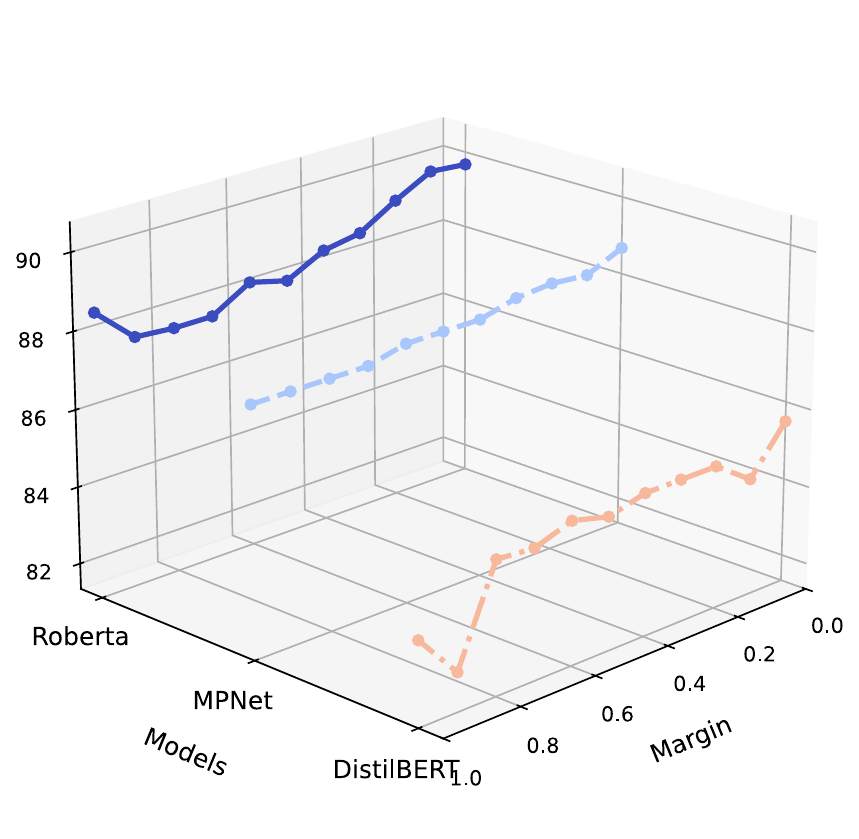}
    \caption{IMDB (Biasing)}
    \label{fig:en_imdb}
  \end{subfigure}
  
  \vspace{0.5em} 

  \begin{subfigure}{0.493\linewidth}
    \centering
    \includegraphics[width=\linewidth]{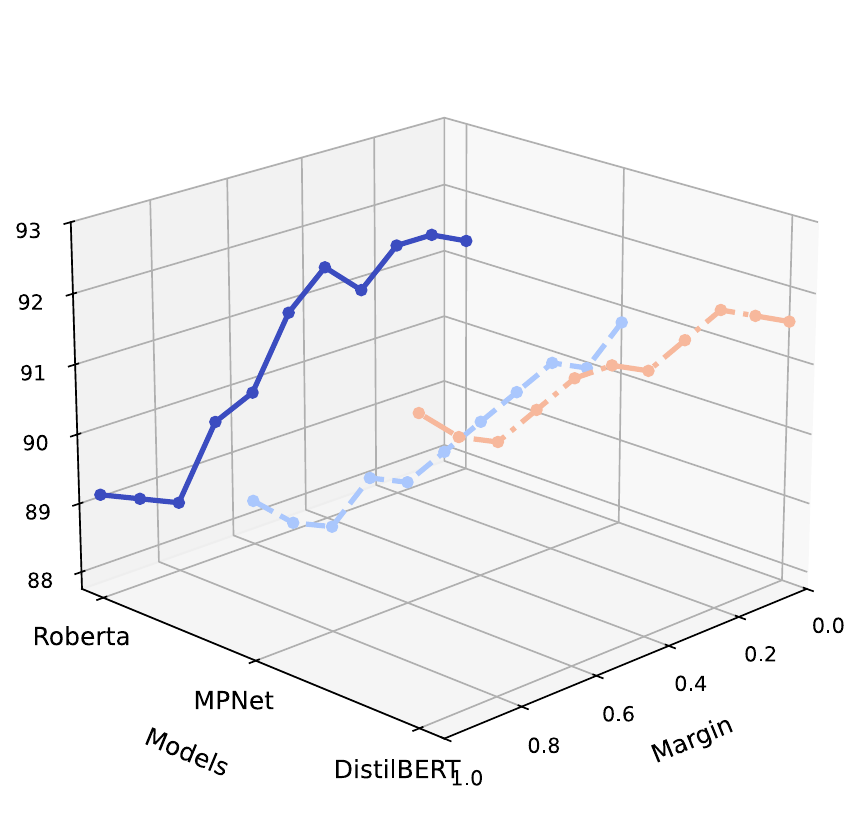}
    \caption{Yelp (Debiasing)}
    \label{fig:de_yelp}
  \end{subfigure}
  \hspace{-0.2em}
  \begin{subfigure}{0.493\linewidth}
    \centering
    \includegraphics[width=\linewidth]{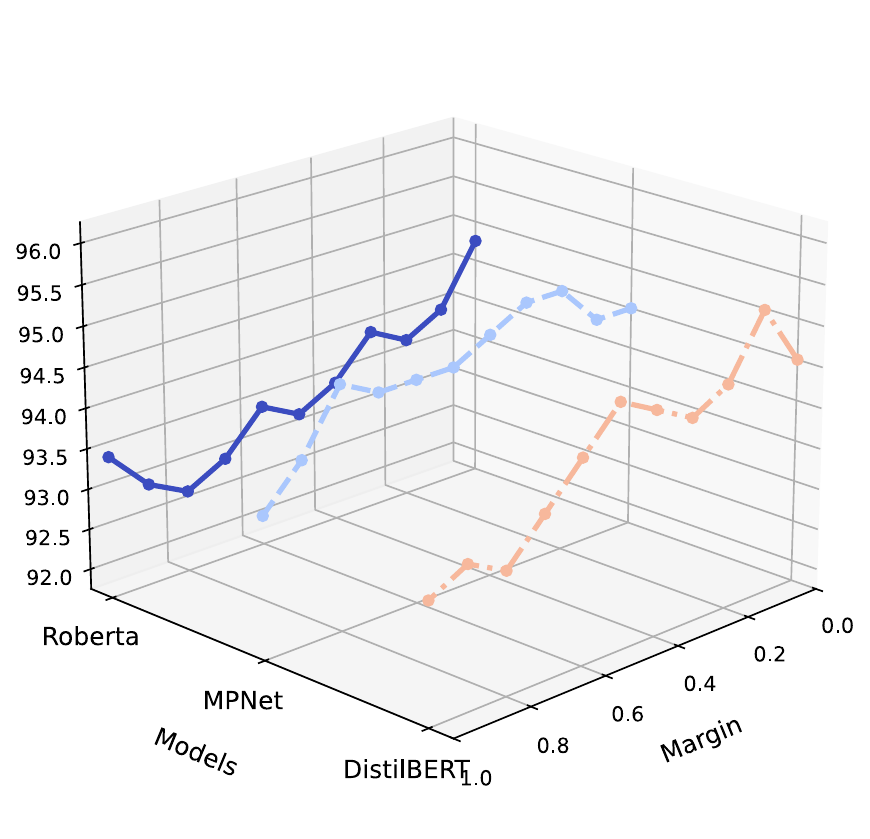}
    \caption{Yelp (Biasing)}
    \label{fig:en_yelp}
  \end{subfigure}

  \caption{The impact of the margin on classification accuracy. Fig.~\ref{fig:de_imdb} and Fig.~\ref{fig:de_yelp} show cases for reducing shortcuts on OOD test. Fig.~\ref{fig:en_imdb} and Fig.~\ref{fig:en_yelp} show cases for enhancing shortcuts on i.i.d. test.}
  \label{fig:margin}
  \vspace{-5pt}
\end{figure}

\paragraph{The Controllability of Shortcuts}
We demonstrated how to weaken or enhance shortcuts by adjusting the value of the margin $M$ in eq.~\eqref{eq:margin1} and eq.~\eqref{eq:margin1}, as shown in Fig.~\ref{fig:margin}. To ensure a fair comparison, all other training parameters were held constant in this experiment.

The margin has a controlling effect on the shortcut learning, as shown in Figure~\ref{fig:margin}. We observed that with the increase of $M$ increases, the performance of all three models on the two datasets exhibits a volatile decline. This suggests that a higher margin makes our method more permissive in enhancing or suppressing shortcut learning, leading to a corresponding decrease in performance on both i.i.d. and OOD data. Therefore, by adjusting $M$, CURE can quantitatively control the impact of shortcut learning on classification, providing a quantifiable benchmark for future debiasing research in theory.

\section{Conclusion}

In this work, we introduced CURE, a novel and lightweight framework for mitigating conceptual shortcuts in pre-trained language models. CURE enables fine-grained control over conceptual bias retention by systematically disentangling concept-relevant and content-relevant representations. It balances robustness and accuracy based on task requirements. Our experiments on IMDB and Yelp datasets demonstrate that CURE significantly improves out-of-distribution robustness, achieving up to 5-point accuracy gains and 10-points F1 gains over baselines while maintaining minimal computational overhead. Notably, CURE reduces training time by an order of magnitude compared to LLM-driven debiasing approaches, making it a scalable and efficient solution. These results highlight CURE, which reveals the potential of unsupervised conceptual debiasing in enhancing the reliability of language models while preserving critical task-relevant features.

\section*{Limitations}

While CURE demonstrates strong performance and computational efficiency, we acknowledge the following limitations.
First, due to computational constraints, we were unable to include large-scale comparisons against debiasing baselines such as RAZOR~\citep{razor} or Focal Loss~\citep{focalloss} on newer model architectures such as LLaMA3-1B \citep{llama3} and Qwen-2.5 \citep{qwen2.5}. While we conducted a preliminary evaluation on LLaMA3-1B to evaluate the generalization ability of CURE (see Appendix~\ref{sec:appendix_additional}), this was limited to comparisons with standard fine-tuned baselines. A comprehensive benchmarking against other debiasing approaches on these models is left to future work.
Second, although CURE itself does not rely on LLM-driven data augmentation during training, we utilized large language models for a one-time concept annotation step during data preprocessing, following prior work~\citep{concept}. This step does not incur additional inference cost and could be replaced with human-annotated concepts in future applications to reduce reliance on external models. However, we did evaluate the plausibility of these annotations through a human study (see Appendix~\ref{sec:appendix_annotation_quality}), confirming their quality for use in downstream evaluations.

Despite these limitations, CURE remains a scalable and adaptable framework for mitigating conceptual biases in NLP models, paving the way for more robust and generalizable language understanding systems.

\bibliography{custom}

\begin{thebibliography}{39}
\providecommand{\natexlab}[1]{#1}

\bibitem[{Chen et~al.(2023)Chen, Gao, Bosselut, Sabharwal, and
  Richardson}]{disco}
Zeming Chen, Qiyue Gao, Antoine Bosselut, Ashish Sabharwal, and Kyle
  Richardson. 2023.
\newblock \href {https://doi.org/10.18653/v1/2023.acl-long.302} {{DISCO}:
  Distilling counterfactuals with large language models}.
\newblock In \emph{Proceedings of the 61st Annual Meeting of the Association
  for Computational Linguistics (Volume 1: Long Papers)}, pages 5514--5528,
  Toronto, Canada. Association for Computational Linguistics.

\bibitem[{Dai et~al.(2019)Dai, Liang, Qiu, and Huang}]{styletransformer}
Ning Dai, Jianze Liang, Xipeng Qiu, and Xuanjing Huang. 2019.
\newblock \href {https://doi.org/10.18653/v1/P19-1601} {Style transformer:
  Unpaired text style transfer without disentangled latent representation}.
\newblock In \emph{Proceedings of the 57th Annual Meeting of the Association
  for Computational Linguistics}, pages 5997--6007, Florence, Italy.
  Association for Computational Linguistics.

\bibitem[{Devlin et~al.(2019)Devlin, Chang, Lee, and Toutanova}]{bert}
Jacob Devlin, Ming-Wei Chang, Kenton Lee, and Kristina Toutanova. 2019.
\newblock \href {https://doi.org/10.18653/v1/N19-1423} {{BERT}: Pre-training of
  deep bidirectional transformers for language understanding}.
\newblock In \emph{Proceedings of the 2019 Conference of the North {A}merican
  Chapter of the Association for Computational Linguistics: Human Language
  Technologies, Volume 1 (Long and Short Papers)}, pages 4171--4186,
  Minneapolis, Minnesota. Association for Computational Linguistics.

\bibitem[{Du et~al.(2022)Du, Yan, Chen, Liu, Zhao, She, Wu, Wang, and
  Qin}]{du2022less}
Yanrui Du, Jing Yan, Yan Chen, Jing Liu, Sendong Zhao, Qiaoqiao She, Hua Wu,
  Haifeng Wang, and Bing Qin. 2022.
\newblock Less learn shortcut: Analyzing and mitigating learning of spurious
  feature-label correlation.
\newblock \emph{arXiv preprint arXiv:2205.12593}.

\bibitem[{Gilardi et~al.(2023)Gilardi, Alizadeh, and
  Kubli}]{Gilardi2023ChatGPTOC}
Fabrizio Gilardi, Meysam Alizadeh, and Ma{\"e}l Kubli. 2023.
\newblock \href {https://api.semanticscholar.org/CorpusID:257766307} {Chatgpt
  outperforms crowd workers for text-annotation tasks}.
\newblock \emph{Proceedings of the National Academy of Sciences of the United
  States of America}, 120.

\bibitem[{He et~al.(2019)He, Zha, and Wang}]{unlearn}
He~He, Sheng Zha, and Haohan Wang. 2019.
\newblock \href {https://doi.org/10.18653/v1/D19-6115} {Unlearn dataset bias in
  natural language inference by fitting the residual}.
\newblock In \emph{Proceedings of the 2nd Workshop on Deep Learning Approaches
  for Low-Resource NLP (DeepLo 2019)}, pages 132--142, Hong Kong, China.
  Association for Computational Linguistics.

\bibitem[{Hur et~al.(2022)Hur, Lee, Oh, Price, Kim, and
  Choi}]{pmlr-v174-hur22a}
Kyunghoon Hur, Jiyoung Lee, Jungwoo Oh, Wesley Price, Younghak Kim, and Edward
  Choi. 2022.
\newblock \href {https://proceedings.mlr.press/v174/hur22a.html} {Unifying
  heterogeneous electronic health records systems via text-based code
  embedding}.
\newblock In \emph{Proceedings of the Conference on Health, Inference, and
  Learning}, volume 174 of \emph{Proceedings of Machine Learning Research},
  pages 183--203. PMLR.

\bibitem[{Jiménez-Sánchez et~al.(2023)Jiménez-Sánchez, Juodelyte,
  Chamberlain, and Cheplygina}]{chest}
Amelia Jiménez-Sánchez, Dovile Juodelyte, Bethany Chamberlain, and Veronika
  Cheplygina. 2023.
\newblock \href {https://doi.org/10.1109/ISBI53787.2023.10230572} {Detecting
  shortcuts in medical images - a case study in chest x-rays}.
\newblock In \emph{2023 IEEE 20th International Symposium on Biomedical Imaging
  (ISBI)}, pages 1--5.

\bibitem[{Kaushik et~al.(2021)Kaushik, Khilji, Sinha, and
  Pakray}]{kaushik-etal-2021-cnlp}
Darsh Kaushik, Abdullah Faiz Ur~Rahman Khilji, Utkarsh Sinha, and Partha
  Pakray. 2021.
\newblock \href {https://doi.org/10.18653/v1/2021.sdp-1.13} {{CNLP}-{NITS} @
  {L}ong{S}umm 2021: {T}ext{R}ank variant for generating long summaries}.
\newblock In \emph{Proceedings of the Second Workshop on Scholarly Document
  Processing}, pages 103--109, Online. Association for Computational
  Linguistics.

\bibitem[{Kaushik and Lipton(2018)}]{much}
Divyansh Kaushik and Zachary~C. Lipton. 2018.
\newblock \href {https://doi.org/10.18653/v1/D18-1546} {How much reading does
  reading comprehension require? a critical investigation of popular
  benchmarks}.
\newblock In \emph{Proceedings of the 2018 Conference on Empirical Methods in
  Natural Language Processing}, pages 5010--5015, Brussels, Belgium.
  Association for Computational Linguistics.

\bibitem[{Kuiper and Martin(1993)}]{humor}
Nicholas~A Kuiper and Rod~A Martin. 1993.
\newblock \href {https://doi.org/doi:10.1515/humr.1993.6.3.251} {Humor and
  self-concept}.
\newblock \emph{Humor}, 6(3):251--270.

\bibitem[{Kumar et~al.(2019)Kumar, Wintner, Smith, and
  Tsvetkov}]{kumar2019topics}
Sachin Kumar, Shuly Wintner, Noah~A Smith, and Yulia Tsvetkov. 2019.
\newblock Topics to avoid: Demoting latent confounds in text classification.
\newblock \emph{arXiv preprint arXiv:1909.00453}.

\bibitem[{Likert(1932)}]{likert1932technique}
Rensis Likert. 1932.
\newblock A technique for the measurement of attitudes.
\newblock \emph{Archives of Psychology}, 140:1--55.

\bibitem[{Lin et~al.(2020)Lin, Goyal, Girshick, He, and Dollár}]{focalloss}
Tsung-Yi Lin, Priya Goyal, Ross Girshick, Kaiming He, and Piotr Dollár. 2020.
\newblock \href {https://doi.org/10.1109/TPAMI.2018.2858826} {Focal loss for
  dense object detection}.
\newblock \emph{IEEE Transactions on Pattern Analysis and Machine
  Intelligence}, 42(2):318--327.

\bibitem[{Loshchilov and Hutter(2019)}]{adamw}
Ilya Loshchilov and Frank Hutter. 2019.
\newblock \href {https://openreview.net/forum?id=Bkg6RiCqY7} {Decoupled weight
  decay regularization}.
\newblock In \emph{International Conference on Learning Representations}.

\bibitem[{Lundberg and Lee(2017)}]{shap}
Scott~M. Lundberg and Su-In Lee. 2017.
\newblock A unified approach to interpreting model predictions.
\newblock In \emph{Proceedings of the 31st International Conference on Neural
  Information Processing Systems}, NIPS'17, page 4768–4777, Red Hook, NY,
  USA. Curran Associates Inc.

\bibitem[{Maas et~al.(2011)Maas, Daly, Pham, Huang, Ng, and Potts}]{imdb}
Andrew~L. Maas, Raymond~E. Daly, Peter~T. Pham, Dan Huang, Andrew~Y. Ng, and
  Christopher Potts. 2011.
\newblock \href {http://www.aclweb.org/anthology/P11-1015} {Learning word
  vectors for sentiment analysis}.
\newblock In \emph{Proceedings of the 49th Annual Meeting of the Association
  for Computational Linguistics: Human Language Technologies}, pages 142--150,
  Portland, Oregon, USA. Association for Computational Linguistics.

\bibitem[{Meta(2024)}]{llama3}
Meta. 2024.
\newblock \href {https://huggingface.co/meta-llama/Llama-3.2-1B} {Llama 3.2-1b
  model card}.
\newblock Accessed: January 23, 2025.

\bibitem[{Ouyang et~al.(2022)Ouyang, Wu, Jiang, Almeida, Wainwright, Mishkin,
  Zhang, Agarwal, Slama, Ray, Schulman, Hilton, Kelton, Miller, Simens, Askell,
  Welinder, Christiano, Leike, and Lowe}]{instructgpt}
Long Ouyang, Jeff Wu, Xu~Jiang, Diogo Almeida, Carroll~L. Wainwright, Pamela
  Mishkin, Chong Zhang, Sandhini Agarwal, Katarina Slama, Alex Ray, John
  Schulman, Jacob Hilton, Fraser Kelton, Luke Miller, Maddie Simens, Amanda
  Askell, Peter Welinder, Paul Christiano, Jan Leike, and Ryan Lowe. 2022.
\newblock Training language models to follow instructions with human feedback.
\newblock In \emph{Proceedings of the 36th International Conference on Neural
  Information Processing Systems}, NIPS '22, Red Hook, NY, USA. Curran
  Associates Inc.

\bibitem[{Radford et~al.(2019)Radford, Wu, Child, Luan, Amodei, Sutskever
  et~al.}]{gpt2}
Alec Radford, Jeffrey Wu, Rewon Child, David Luan, Dario Amodei, Ilya
  Sutskever, et~al. 2019.
\newblock \href {https://api.semanticscholar.org/CorpusID:160025533} {Language
  models are unsupervised multitask learners}.
\newblock \emph{OpenAI blog}, 1(8):9.

\bibitem[{Sagawa* et~al.(2020)Sagawa*, Koh*, Hashimoto, and
  Liang}]{Sagawa*2020Distributionally}
Shiori Sagawa*, Pang~Wei Koh*, Tatsunori~B. Hashimoto, and Percy Liang. 2020.
\newblock \href {https://openreview.net/forum?id=ryxGuJrFvS} {Distributionally
  robust neural networks}.
\newblock In \emph{International Conference on Learning Representations}.

\bibitem[{Sennrich et~al.(2016)Sennrich, Haddow, and Birch}]{backtrans}
Rico Sennrich, Barry Haddow, and Alexandra Birch. 2016.
\newblock \href {https://doi.org/10.18653/v1/P16-1009} {Improving neural
  machine translation models with monolingual data}.
\newblock In \emph{Proceedings of the 54th Annual Meeting of the Association
  for Computational Linguistics (Volume 1: Long Papers)}, pages 86--96, Berlin,
  Germany. Association for Computational Linguistics.

\bibitem[{Shazeer(2020)}]{swiglu}
Noam~M. Shazeer. 2020.
\newblock \href {https://api.semanticscholar.org/CorpusID:211096588} {Glu
  variants improve transformer}.
\newblock \emph{ArXiv}, abs/2002.05202.

\bibitem[{Stacey et~al.(2020)Stacey, Minervini, Dubossarsky, Riedel, and
  Rockt{\"a}schel}]{stacey2020avoiding}
Joe Stacey, Pasquale Minervini, Haim Dubossarsky, Sebastian Riedel, and Tim
  Rockt{\"a}schel. 2020.
\newblock Avoiding the hypothesis-only bias in natural language inference via
  ensemble adversarial training.
\newblock \emph{arXiv preprint arXiv:2004.07790}.

\bibitem[{Tishby et~al.(1999)Tishby, Pereira, and Bialek}]{tishby99information}
Naftali Tishby, Fernando~C. Pereira, and William Bialek. 1999.
\newblock \href {https://arxiv.org/abs/physics/0004057} {The information
  bottleneck method}.
\newblock In \emph{Proc. of the 37-th Annual Allerton Conference on
  Communication, Control and Computing}, pages 368--377.

\bibitem[{Touvron et~al.(2023{\natexlab{a}})Touvron, Lavril, Izacard, Martinet,
  Lachaux, Lacroix, Rozière, Goyal, Hambro, Azhar, Rodriguez, Joulin, Grave,
  and Lample}]{llama}
Hugo Touvron, Thibaut Lavril, Gautier Izacard, Xavier Martinet, Marie-Anne
  Lachaux, Timothée Lacroix, Baptiste Rozière, Naman Goyal, Eric Hambro,
  Faisal Azhar, Aurelien Rodriguez, Armand Joulin, Edouard Grave, and Guillaume
  Lample. 2023{\natexlab{a}}.
\newblock \href {https://arxiv.org/abs/2302.13971} {Llama: Open and efficient
  foundation language models}.
\newblock \emph{Preprint}, arXiv:2302.13971.

\bibitem[{Touvron et~al.(2023{\natexlab{b}})Touvron, Martin, Stone, Albert,
  Almahairi, Babaei, Bashlykov, Batra, Bhargava, Bhosale, Bikel, Blecher,
  Ferrer, Chen, Cucurull, Esiobu, Fernandes, Fu, Fu, Fuller, Gao, Goswami,
  Goyal, Hartshorn, Hosseini, Hou, Inan, Kardas, Kerkez, Khabsa, Kloumann,
  Korenev, Koura, Lachaux, Lavril, Lee, Liskovich, Lu, Mao, Martinet, Mihaylov,
  Mishra, Molybog, Nie, Poulton, Reizenstein, Rungta, Saladi, Schelten, Silva,
  Smith, Subramanian, Tan, Tang, Taylor, Williams, Kuan, Xu, Yan, Zarov, Zhang,
  Fan, Kambadur, Narang, Rodriguez, Stojnic, Edunov, and Scialom}]{llama2}
Hugo Touvron, Louis Martin, Kevin Stone, Peter Albert, Amjad Almahairi, Yasmine
  Babaei, Nikolay Bashlykov, Soumya Batra, Prajjwal Bhargava, Shruti Bhosale,
  Dan Bikel, Lukas Blecher, Cristian~Canton Ferrer, Moya Chen, Guillem
  Cucurull, David Esiobu, Jude Fernandes, Jeremy Fu, Wenyin Fu, Brian Fuller,
  Cynthia Gao, Vedanuj Goswami, Naman Goyal, Anthony Hartshorn, Saghar
  Hosseini, Rui Hou, Hakan Inan, Marcin Kardas, Viktor Kerkez, Madian Khabsa,
  Isabel Kloumann, Artem Korenev, Punit~Singh Koura, Marie-Anne Lachaux,
  Thibaut Lavril, Jenya Lee, Diana Liskovich, Yinghai Lu, Yuning Mao, Xavier
  Martinet, Todor Mihaylov, Pushkar Mishra, Igor Molybog, Yixin Nie, Andrew
  Poulton, Jeremy Reizenstein, Rashi Rungta, Kalyan Saladi, Alan Schelten, Ruan
  Silva, Eric~Michael Smith, Ranjan Subramanian, Xiaoqing~Ellen Tan, Binh Tang,
  Ross Taylor, Adina Williams, Jian~Xiang Kuan, Puxin Xu, Zheng Yan, Iliyan
  Zarov, Yuchen Zhang, Angela Fan, Melanie Kambadur, Sharan Narang, Aurelien
  Rodriguez, Robert Stojnic, Sergey Edunov, and Thomas Scialom.
  2023{\natexlab{b}}.
\newblock \href {https://arxiv.org/abs/2307.09288} {Llama 2: Open foundation
  and fine-tuned chat models}.
\newblock \emph{Preprint}, arXiv:2307.09288.

\bibitem[{Tu et~al.(2020)Tu, Lalwani, Gella, and He}]{tu2020empirical}
Lifu Tu, Garima Lalwani, Spandana Gella, and He~He. 2020.
\newblock An empirical study on robustness to spurious correlations using
  pre-trained language models.
\newblock \emph{Transactions of the Association for Computational Linguistics},
  8:621--633.

\bibitem[{Vaswani et~al.(2017)Vaswani, Shazeer, Parmar, Uszkoreit, Jones,
  Gomez, Kaiser, and Polosukhin}]{attention}
Ashish Vaswani, Noam Shazeer, Niki Parmar, Jakob Uszkoreit, Llion Jones,
  Aidan~N Gomez, \L~ukasz Kaiser, and Illia Polosukhin. 2017.
\newblock \href
  {https://proceedings.neurips.cc/paper_files/paper/2017/file/3f5ee243547dee91fbd053c1c4a845aa-Paper.pdf}
  {Attention is all you need}.
\newblock In \emph{Advances in Neural Information Processing Systems},
  volume~30. Curran Associates, Inc.

\bibitem[{Wang et~al.(2022)Wang, Zhou, Sun, Ye, Gui, Zhang, and
  Huang}]{wang-etal-2022-causal}
Siyin Wang, Jie Zhou, Changzhi Sun, Junjie Ye, Tao Gui, Qi~Zhang, and Xuanjing
  Huang. 2022.
\newblock \href {https://aclanthology.org/2022.coling-1.607} {Causal
  intervention improves implicit sentiment analysis}.
\newblock In \emph{Proceedings of the 29th International Conference on
  Computational Linguistics}, pages 6966--6977, Gyeongju, Republic of Korea.
  International Committee on Computational Linguistics.

\bibitem[{Wang et~al.(2021)Wang, Sridhar, Yang, and Wang}]{wang2021identifying}
Tianlu Wang, Rohit Sridhar, Diyi Yang, and Xuezhi Wang. 2021.
\newblock Identifying and mitigating spurious correlations for improving
  robustness in nlp models.
\newblock \emph{arXiv preprint arXiv:2110.07736}.

\bibitem[{Wang and Culotta(2020)}]{wang2020identifying}
Zhao Wang and Aron Culotta. 2020.
\newblock Identifying spurious correlations for robust text classification.
\newblock \emph{arXiv preprint arXiv:2010.02458}.

\bibitem[{Wen et~al.(2022)Wen, Zhu, Zhang, Zhou, and Huang}]{cad}
Jiaxin Wen, Yeshuang Zhu, Jinchao Zhang, Jie Zhou, and Minlie Huang. 2022.
\newblock \href {https://doi.org/10.18653/v1/2022.findings-emnlp.170}
  {{A}uto{CAD}: Automatically generate counterfactuals for mitigating shortcut
  learning}.
\newblock In \emph{Findings of the Association for Computational Linguistics:
  EMNLP 2022}, pages 2302--2317, Abu Dhabi, United Arab Emirates. Association
  for Computational Linguistics.

\bibitem[{Xu et~al.(2023)Xu, Liu, Wu, and Wang}]{fever-cf}
Weizhi Xu, Qiang Liu, Shu Wu, and Liang Wang. 2023.
\newblock \href {https://doi.org/10.18653/v1/2023.acl-long.374} {Counterfactual
  debiasing for fact verification}.
\newblock In \emph{Proceedings of the 61st Annual Meeting of the Association
  for Computational Linguistics (Volume 1: Long Papers)}, pages 6777--6789,
  Toronto, Canada. Association for Computational Linguistics.

\bibitem[{Yaghoobzadeh et~al.(2019)Yaghoobzadeh, Mehri, Tachet, Hazen, and
  Sordoni}]{yaghoobzadeh2019increasing}
Yadollah Yaghoobzadeh, Soroush Mehri, Remi Tachet, Timothy~J Hazen, and
  Alessandro Sordoni. 2019.
\newblock Increasing robustness to spurious correlations using forgettable
  examples.
\newblock \emph{arXiv preprint arXiv:1911.03861}.

\bibitem[{Yang et~al.(2024{\natexlab{a}})Yang, Yang, Zhang, Hui, Zheng, Yu, Li,
  Liu, Huang, Wei, Lin, Yang, Tu, Zhang, Yang, Yang, Zhou, Lin, Dang, Lu, Bao,
  Yang, Yu, Li, Xue, Zhang, Zhu, Men, Lin, Li, Xia, Ren, Ren, Fan, Su, Zhang,
  Wan, Liu, Cui, Zhang, and Qiu}]{qwen2.5}
An~Yang, Baosong Yang, Beichen Zhang, Binyuan Hui, Bo~Zheng, Bowen Yu,
  Chengyuan Li, Dayiheng Liu, Fei Huang, Haoran Wei, Huan Lin, Jian Yang,
  Jianhong Tu, Jianwei Zhang, Jianxin Yang, Jiaxi Yang, Jingren Zhou, Junyang
  Lin, Kai Dang, Keming Lu, Keqin Bao, Kexin Yang, Le~Yu, Mei Li, Mingfeng Xue,
  Pei Zhang, Qin Zhu, Rui Men, Runji Lin, Tianhao Li, Tingyu Xia, Xingzhang
  Ren, Xuancheng Ren, Yang Fan, Yang Su, Yichang Zhang, Yu~Wan, Yuqiong Liu,
  Zeyu Cui, Zhenru Zhang, and Zihan Qiu. 2024{\natexlab{a}}.
\newblock Qwen2.5 technical report.
\newblock \emph{arXiv preprint arXiv:2412.15115}.

\bibitem[{Yang et~al.(2024{\natexlab{b}})Yang, Prenkaj, and Kasneci}]{razor}
Shuo Yang, Bardh Prenkaj, and Gjergji Kasneci. 2024{\natexlab{b}}.
\newblock \href {https://arxiv.org/abs/2412.07675} {Razor: Sharpening knowledge
  by cutting bias with unsupervised text rewriting}.
\newblock \emph{Preprint}, arXiv:2412.07675.

\bibitem[{Zhang et~al.(2015)Zhang, Zhao, and LeCun}]{yelp}
Xiang Zhang, Junbo Zhao, and Yann LeCun. 2015.
\newblock Character-level convolutional networks for text classification.
\newblock In \emph{Proceedings of the 29th International Conference on Neural
  Information Processing Systems - Volume 1}, NIPS'15, page 649–657,
  Cambridge, MA, USA. MIT Press.

\bibitem[{Zhou et~al.(2024)Zhou, Xu, Liu, An, Ai, and Huang}]{concept}
Yuhang Zhou, Paiheng Xu, Xiaoyu Liu, Bang An, Wei Ai, and Furong Huang. 2024.
\newblock \href {https://doi.org/10.18653/v1/2024.acl-long.28} {Explore
  spurious correlations at the concept level in language models for text
  classification}.
\newblock In \emph{Proceedings of the 62nd Annual Meeting of the Association
  for Computational Linguistics (Volume 1: Long Papers)}, pages 478--492,
  Bangkok, Thailand. Association for Computational Linguistics.

\end{thebibliography}

\onecolumn
\appendix
\section{Appendix}
\subsection{Training Algorithm}
\begin{algorithm}
    \caption{The Training Algorithm of CURE}
    \label{al:main}
    \begin{algorithmic}[1]
        \Require A biased dataset $D$, a corresponding label set $\mathcal{Y}$, a corresponding concept set $\mathcal{C}$.
        \Ensure A robust classifier $\theta$.
        \State Train a concept classifier $\omega$ using \eqref{eq:concept}.
        \For{$d \in D$}
            \State Input $d$ to a PLM, obtain output $x$.
            \State Freeze a content extractor $\phi$ and train a reversal extractor $\hat{\phi}$ using \eqref{eq:reconstruction_loss}.
            \State Freeze $\hat{\phi}$ and train $\phi$ using \eqref{eq:phi}.
        \EndFor
        
        \For{$d \in D$}
            \State Freeze $\phi$ and train a debiasing module $\psi$ using \eqref{eq:margin1}.
        \EndFor
        \State Train a task classifier $\theta$ using \eqref{eq:ce}.
    \end{algorithmic}
\end{algorithm}

\clearpage

\subsection{Prompt Design for Concept Labeling}\label{prompt}

\afterpage{\twocolumn}
\noindent
\begin{table*}[ht]
\centering
\resizebox{\textwidth}{!}{
\begin{tabular}{p{16cm}}
\toprule
\noindent
Here is a given movie review: 
\vspace{0.2cm}

\textbf{\{review\}} 
\vspace{0.2cm}

Identify the main concept discussed in this review using only ONE WORD. Your response should be ONE-WORD for each review (e.g., acting, plot, cinematography). 
\vspace{0.2cm}

\noindent
Examples: \\

1. Review: 
``Seen  `Back to the Future'? This movie, `Tangents' (aka 'Time Chasers'), tries a similar time-travel concept but fails to hit the mark. Made in 1994, it looks and feels like it’s from the 80s. The cast includes an unappealing leading man, a cliché-ridden leading lady, a cartoonish villain, and henchmen with questionable jobs. The plot is hard to follow, so I’d recommend watching it with Mystery Science Theater 3000 for entertainment. On its own: 3 stars. With MST3K: 8 stars.'' \\
Concept: plot 
\vspace{0.2cm}

2. Review:
``And you thought your significant other's family was weird? Wedding Slashers will make you think twice about ever saying `I do.' It is reminiscent of past horror titles such as `Deadly Friend' and `Friday the 13th.' It is a classic slasher film that features characters with names like `Sock Monkey' and `The Mortician.' You may laugh at first but trust me, these guys will freak you out. This is a quencher for the blood-thirsty horror/slasher fan that needs to see gore, gore and more gore. It's not all slash and gash either - Wedding Slashers is chock-full-of one-liners and will give you more than just a chuckle. You're going to need to see this one to believe it.''  \\
Concept: genre  

\vspace{0.3cm}
\noindent
Now, classify the given review and provide the main concept using only ONE WORD: \\

\bottomrule
\end{tabular}}
\label{tab:prompt_concept}
\caption{Prompt $P_a$ is used to label the IMDB dataset for concept annotation. The placeholder \textbf{\{review\}} represents the input movie review. The example reviews are also sourced from the IMDB dataset.}
\vspace{-10pt}
\end{table*}

\vspace{0.8 cm}

\begin{table*}[h]
\centering
\resizebox{\textwidth}{!}{
\begin{tabular}{p{16cm}}
\toprule
\noindent
Here is a list of extracted concepts from movie reviews: \textbf{\{concepts\}} \\
Analyze these concepts and suggest an appropriate number of clusters and one-word cluster names to group them. Cluster names should not overlap, should be distinctive.
\\
\bottomrule
\end{tabular}
}
\caption{Prompt $P_b$ is used to refine the concept clusters, and it returns final concept list $\mathcal{C}$. The placeholder \textbf{\{concepts\}} represents the concepts that are generated using $P_a$.}
\vspace{-5pt}
\end{table*}

\vspace{0.8 cm}

\begin{table*}[h]
\centering
\resizebox{\textwidth}{!}{
\begin{tabular}{p{16cm}}
\toprule
\noindent
Given concept: \textbf{\{concept\}} \\
Predefined Concept List: \textbf{\{concept labels\}} \\
Provide the concept from the predefined list that is closest to the given concept. Return nothing else.
\\
\bottomrule
\end{tabular}}
\caption{Prompt $P_c$ is used to assign the final concept from $\mathcal{C}$ to each movie review. The placeholder \textbf{\{concept\}} represents the extracted concept from a movie review. The placeholder \textbf{\{concept labels\}} refers to $\mathcal{C}$, the predefined concept list generated using $P_b$.}
\end{table*}

\clearpage

\onecolumn
\subsection{Annotation Quality Evaluation}
\label{sec:appendix_annotation_quality}

To measure the quality of GPT-4o's concept annotations, we conducted a human evaluation using crowdsourcing. We randomly selected 10 annotations from each dataset (Yelp and IMDB), and each annotation was rated by seven independent annotators using Qualtrics \footnote{\url{https://www.qualtrics.com/}}. The annotators assessed how accurately each concept reflected the associated text using a 5-point Likert scale~\cite{likert1932technique}, where 1 = \textit{Not accurately at all}, 2 = \textit{Slightly accurately}, 3 = \textit{Moderately accurately}, 4 = \textit{Very accurately}, and 5 = \textit{Extremely accurately}. The average ratings were 4.31 for Yelp and 3.81 for IMDB. We define the \textit{agreement rate} as the proportion of ratings above 3, which reached 100\% for Yelp and 70\% for IMDB. These results indicate that GPT-4o’s concept annotations are largely considered plausible and can be reliably used in downstream tasks.

\vspace{0.5 cm}

\begin{table}[ht]
\centering
\begin{tabular}{@{}lcc@{}}
\toprule
\textbf{Dataset} & \textbf{Mean Rating} & \textbf{Agreement Rate (\%)} \\
\midrule
Yelp & 4.31 & 100 \\
IMDB & 3.81 & 70 \\
\bottomrule
\end{tabular}
\caption{Human evaluation of GPT-4o’s concept annotations.}
\end{table}

\vspace{1 cm}


\subsection{Supplementary Experiments}
\label{sec:appendix_additional}

To further demonstrate the generalization capability of our method, we apply CURE to the LLaMA3-1B model~\citep{llama3} to evaluate its effectiveness on a recent large-scale foundation model. Using the same evaluation protocol, we observe a 3-point accuracy improvement on IMDB and consistent gains on Yelp (Table~\ref{tab:llama3}), reflecting the performance trends previously seen with smaller PLMs (Table~\ref{tab:main}). Due to computational resource limitations, this additional experiment could not be extended to include comparisons with other baselines such as RAZOR \cite{razor} and FL \cite{focalloss}, and was therefore not included in the main body of the study. Nonetheless, we report it here in the appendix to highlight the broader applicability and robustness of CURE across diverse and emerging model architectures.

\vspace{0.5 cm}

\begin{table*}[h]
\renewcommand{\arraystretch}{1.5} 
\centering
\small
\begin{tabular}{lllllllllllllll}
\toprule
\multicolumn{2}{c}{\textbf{Dataset}}  & \multicolumn{6}{c}{\textbf{IMDB}}    & \multicolumn{1}{c}{}     & \multicolumn{6}{c}{\textbf{Yelp}}     \\ \cline{3-8} \cline{10-15} 
\multicolumn{2}{c}{\textbf{Model}}    & \multicolumn{2}{c}{ACC $\uparrow$} & \multicolumn{2}{c}{F1 $\uparrow$} & \multicolumn{2}{c}{} & \multicolumn{1}{c}{}     & \multicolumn{2}{c}{ACC $\uparrow$} & \multicolumn{2}{c}{F1 $\uparrow$} & \multicolumn{2}{c}{} \\ \hline
                             & LLaMA3-1B             & \multicolumn{2}{c}{70.67} & \multicolumn{2}{c}{65.00} & \multicolumn{2}{c}{} &                          & \multicolumn{2}{c}{93.25} & \multicolumn{2}{c}{93.00} & \multicolumn{2}{c}{} \\
                             & \cellcolor[HTML]{EFEFEF}LLaMA3-1B (w/ CURE) & \multicolumn{2}{c}{\cellcolor[HTML]{EFEFEF}\textbf{73.83}} & \multicolumn{2}{c}{\cellcolor[HTML]{EFEFEF}\textbf{75.00}} & \multicolumn{2}{c}{\cellcolor[HTML]{EFEFEF}} & \cellcolor[HTML]{EFEFEF} & \multicolumn{2}{c}{\cellcolor[HTML]{EFEFEF}\textbf{93.50}} & \multicolumn{2}{c}{\cellcolor[HTML]{EFEFEF}\textbf{94.00}} & \multicolumn{2}{c}{\cellcolor[HTML]{EFEFEF}} \\
\bottomrule
\end{tabular}
\caption{Accuracy and F1 results on IMDB and Yelp using LLaMA3-1B. ``LLaMA3-1B'' refers to the base model without debiasing. \textbf{Bold} indicates the best result.}
\label{tab:llama3}
\vspace{-5pt}
\end{table*}

\newpage

\subsection{Sentiment Distributions in the Imbalanced Groups}

\begin{figure}[ht]
  \centering
  
  \begin{subfigure}{0.49\linewidth}
    \centering
    \includegraphics[width=\linewidth]{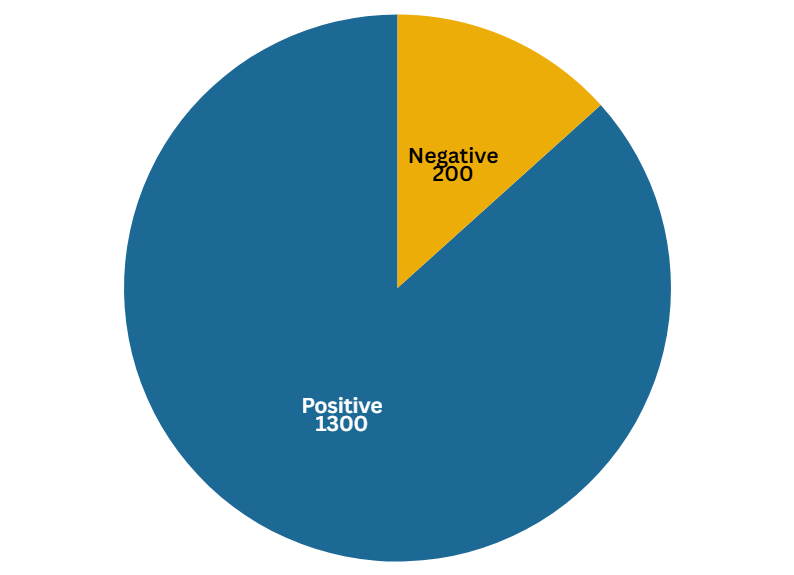}
    \caption{Sentiment distribution in the concept ``Emotion''}
    \label{fig:imdb_emotion}
  \end{subfigure}
  \hspace{-0.3em}
  \begin{subfigure}{0.49\linewidth}
    \centering
    \includegraphics[width=\linewidth]{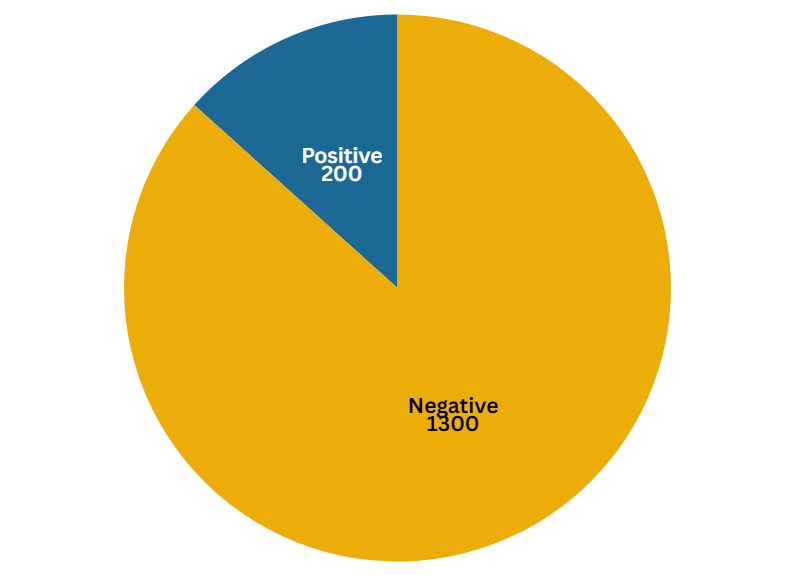}
    \caption{Sentiment distribution in the concept ``Story''}
    \label{fig:imdb_story}
  \end{subfigure}
  
  \vspace{0.6em} 
  
  \begin{subfigure}{0.49\linewidth}
    \centering
    \includegraphics[width=\linewidth]{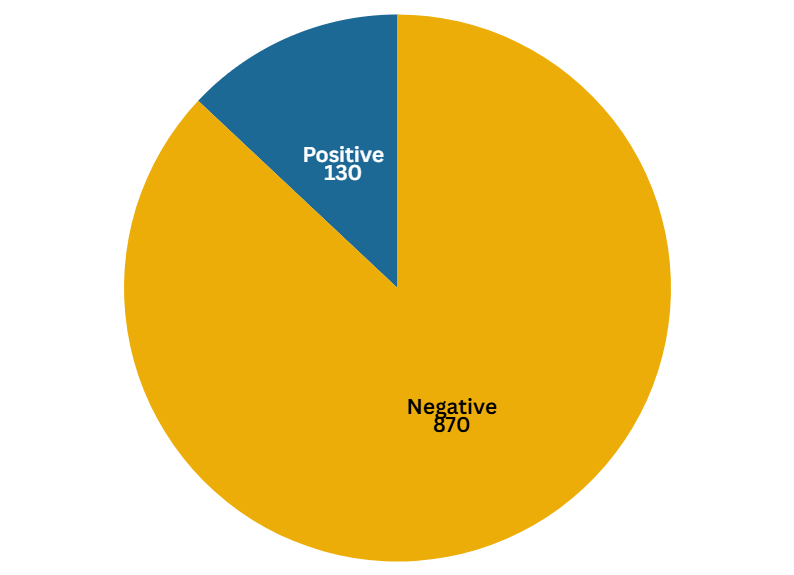}
    \caption{Sentiment distribution in the concept ``Experience''}
    \label{fig:yelp_exp}
  \end{subfigure}
  \hspace{-0.3em}
  \begin{subfigure}{0.49\linewidth}
    \centering
    \includegraphics[width=\linewidth]{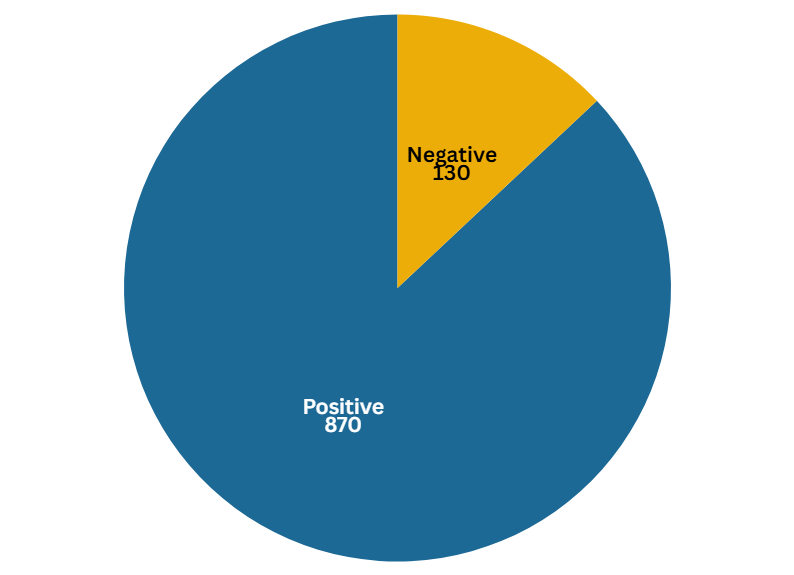}
    \caption{Sentiment distribution in the concept ``Service''}
    \label{fig:yelp_service}
  \end{subfigure}
  
  \caption{Sentiment distributions in the imbalanced groups of the IMDB and Yelp datasets.}
  \label{fig:imbalance}
  \vspace{-8pt}
\end{figure}

\end{document}